\documentclass[11pt]{article}

\usepackage{acl}

\usepackage{times}
\usepackage{latexsym}
\usepackage{enumitem}
\usepackage[T1]{fontenc}
\usepackage{amssymb}
\usepackage{amsmath}
\usepackage{makecell}
\usepackage[utf8]{inputenc}
\usepackage{amsmath}
\usepackage{microtype}

\usepackage{inconsolata}

\usepackage{graphicx}
\usepackage{booktabs}
\usepackage{multirow}

\title{Beyond Accuracy: Evaluating Grounded Visual Evidence in Thinking with Images}

\author{\textbf{Xuchen Li}$^{1}$\thanks{Equal contribution.} \hspace{9pt} 
\textbf{Xuzhao Li}$^{2*}$\hspace{9pt}
\textbf{Renjie Pi}$^{3}$\thanks{Project Leader.}\hspace{9pt}
\textbf{Shiyu Hu}$^{2}$\hspace{9pt}
\textbf{Jian Zhao}$^{1,5}$\hspace{9pt}
\textbf{Jiahui Gao}$^{4}$\thanks{Corresponding Author.} \hspace{9pt}\\
\textsuperscript{1}ZGCA,
\textsuperscript{2}NTU,
\textsuperscript{3}HKUST,
\textsuperscript{4}HKU,
\textsuperscript{5}ZGCI\\
\tt\small xuzhaoli2001@gmail.com, xuchenli1030@gmail.com, rpi@connect.ust.hk, ggaojiahui@gmail.com
}

\begin{document}
\maketitle
\begin{abstract}
Despite the remarkable progress of Vision-Language Models (VLMs) in adopting "Thinking-with-Images" capabilities, accurately evaluating the authenticity of their reasoning process remains a critical challenge.
Existing benchmarks mainly rely on outcome-oriented accuracy, lacking the capability to assess whether models can accurately leverage fine-grained visual cues for multi-step reasoning. 
To address these limitations, we propose ViEBench, a process-verifiable benchmark designed to evaluate faithful visual reasoning.
Comprising 200 multi-scenario high-resolution images with expert-annotated visual evidence, ViEBench uniquely categorizes tasks by difficulty into \textit{perception} and \textit{reasoning} dimensions, where reasoning tasks require utilizing localized visual details with prior knowledge.
To establish comprehensive evaluation criteria, we introduce a dual-axis matrix that provides fine-grained metrics through four diagnostic quadrants, enabling transparent diagnosis of model behavior across varying task complexities.
Our experiments yield several interesting observations: (1) VLMs can sometimes produce correct final answers despite grounding on irrelevant regions, and (2) they may successfully locate the correct evidence but still fail to utilize it to reach accurate conclusions.
Our findings demonstrate that ViEBench can serve as a more explainable and practical benchmark for comprehensively evaluating the effectiveness agentic VLMs. The codes will be released at: https://github.com/Xuchen-Li/ViEBench.
\end{abstract}

\section{Introduction}
Recent advancements in Vision-Language Models (VLMs) \cite{qwen3vl,pixelreasoner,minio3,kimi-k1.5,kimi-k2,minimax-m1} have 
entered
in a new era of "Thinking-with-Images," 
where models no longer perceive images as static inputs but instead actively interact with them through dynamic visual operations~\cite{deepeyes}.
By adopting agentic behaviors such as autonomous zooming, state-of-the-art agentic models like the o3 \cite{o3} and Qwen3-VL \cite{qwen3vl} have demonstrated an unprecedented ability to resolve fine-grained details within high-resolution scenes. This shift from passive perception to active visual reasoning has enabled VLMs to tackle complex tasks in real-world scenarios, ranging from industrial inspection to urban navigation \cite{su2025thinking,zhang2023unmanned,li2025select,cao2025large}.

\begin{table}[t]
\centering
\caption{Comparison of ViEBench with existing Multimodal Benchmarks. ViEBench is uniquely categorized into perception and reasoning tasks, and provides expert-annotated bounding box (BBox) for visual evidence, enabling a precise process evaluation.}
\vspace{-5pt}
\label{tab:benchmark_comparison}
\resizebox{0.48\textwidth}{!}{
\setlength{\tabcolsep}{2.5pt}
\begin{tabular}{lcccccc}
\toprule
\textbf{Bench} & \textbf{\#QA Pairs} & \textbf{Percept} & \textbf{Reason} & \textbf{BBox} & \textbf{\makecell{Process\\Evaluation}} \\
\midrule
V* Bench     & 191 & \checkmark & $\times$ & $\times$ & $\times$ \\
HRBench    & 1600   & \checkmark & $\times$ & $\times$ & $\times$ \\
InfoVQA    & 2801  & \checkmark & $\times$ & $\times$ & $\times$ \\
VisualProbe      & 515    & \checkmark & $\times$ & $\times$ & $\times$ \\
\midrule
\textbf{ViEBench} & \textbf{200} & \checkmark & \checkmark & \checkmark & \checkmark \\
\bottomrule
\end{tabular}
}
\vspace{-15pt}
\end{table}

However, as VLMs gain the autonomy to manipulate their own visual input, a critical evaluation gap has emerged. As shown in Tab. \ref{tab:benchmark_comparison}, existing benchmarks \cite{hrbench,seal,minio3,infovqa,chen2024we} are constrained by two fundamental limitations. First, their perception-oriented tasks can be addressed through fine-grained recognition alone, making them insufficient for evaluating tool-use capabilities in reasoning-intensive scenarios where models must integrate localized visual cues with prior knowledge. Second, these benchmarks rely solely on outcome-oriented metrics \cite{li2025causalstep,li2025sciagent}, treating models as "black boxes" and providing no diagnostic granularity to distinguish whether performance degradation results from grounding failures or from an inability to synthesize visual evidence into logical reasoning. Consequently, without fine-grained metrics to verify if a model's success relies on faithful reasoning or mere textual priors, it remains impossible to pinpoint specific weaknesses or guide targeted model improvements.

\begin{figure*}[htbp!]
    \centering
    \includegraphics[width=\textwidth]{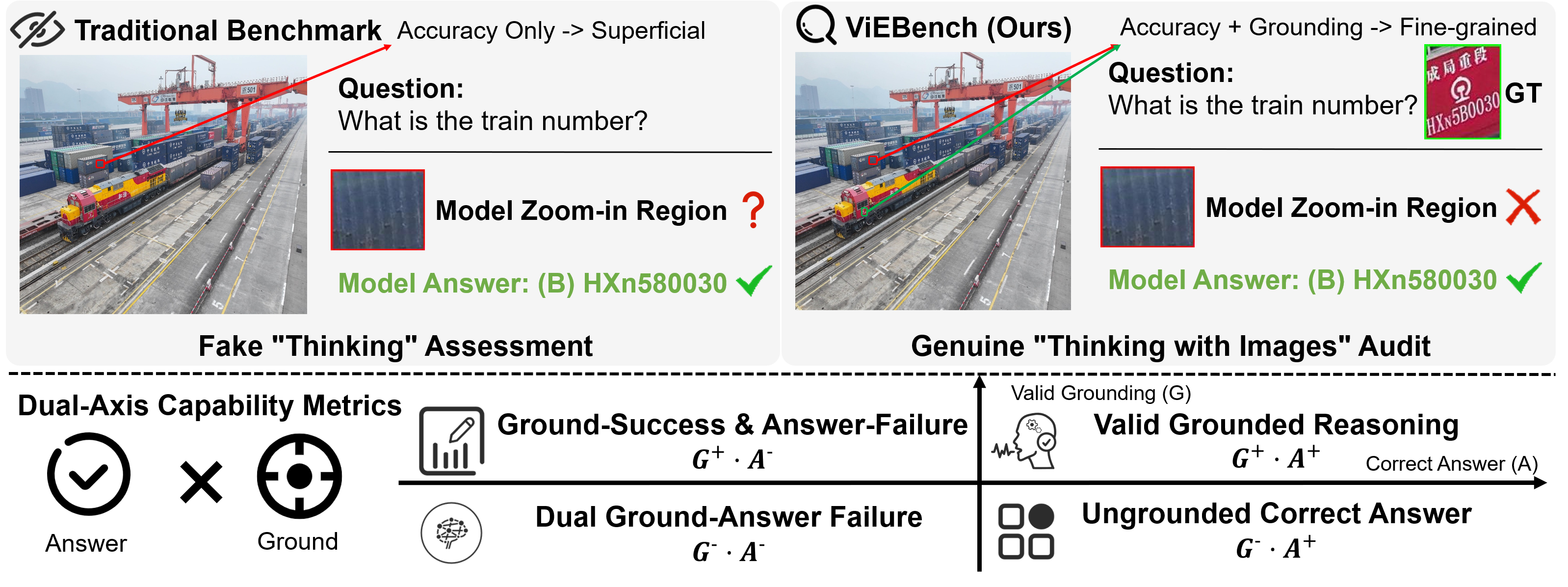}
    \vspace{-15pt}
    \caption{Traditional benchmarks provide a superficial thinking evaluation by relying solely on final answer accuracy, which fails to detect if a model correctly answers via irrelevant visual regions. In contrast, ViEBench performs a faithful "Thinking-with-Images" audit by cross-referencing answer accuracy with visual grounding alignment. Our dual-axis capability matrix deconstructs VLMs performance into four fine-grained quadrants to provide a diagnostic map that identifies whether correct predictions are rooted in sound visual evidence.}
    \label{fig:insight}
    \vspace{-15pt}
\end{figure*}

To bridge this gap, we introduce ViEBench, a novel diagnostic benchmark to evaluate the "Thinking-with-Images" capabilities of VLMs. 
First, we design \textbf{ViEBench-R}, a reasoning task that requires models to localize fine-grained visual cues, integrate them with prior knowledge, and perform multi-step logical reasoning. We also provide \textbf{ViEBench-P}, a perception task for comparable analysis.
To ensure task quality and diversity, we curate 200 high-resolution images across four critical real-world scenarios: retail, urban, industry, and daily life. Crucially, we enforce extreme spatial sparsity, where critical visual evidence occupies less than 0.7\% of the total image area on average. This strategic design ensures that essential visual cues remain sub-perceptual in global views, forcing models to execute precise local zooming operations.

Second, we introduce a dual-axis capability matrix based on the intersection over area (IoA) metric \cite{ioa}. By comparing model-generated visual crops against expert-annotated gold BBox, we construct a grounding axis that quantifies localization accuracy, which is then crossed with the answer correctness axis to form four diagnostic quadrants (shown in Fig.~\ref{fig:insight}): \textit{Valid Grounded Reasoning},  \textit{Ground-Success Answer-Failure},  \textit{Ungrounded Correct Answer}, and \textit{Dual Ground-Answer Failure}. This decomposition enables us to evaluate whether performance degradation stems from grounding failures or reasoning deficiencies—a diagnostic capability completely absent in traditional accuracy-only metrics.

Our extensive evaluation of both end-to-end and agentic VLMs reveals several counter-intuitive findings. We identify a prevalent ungrounded correct answer phenomenon among agentic VLMs, suggesting that current benchmarks significantly overestimate model reliability. Furthermore, we uncover a ground-success and answer-failure bottleneck, where models successfully locate the required evidence but fail to synthesize it into a correct reasoning chain. These insights underscore that the next frontier for VLMs lies not just in "where to look," but in how to deeply integrate cropped visual information into the reasoning chain. By providing a rigorous and transparent evaluation protocol, ViEBench serves as both a valuable benchmark for current models and a roadmap for the development of more robust and truly visual "thinking" agentic models.

This work makes three key contributions:
\textbf{1) }We introduce ViEBench across diverse real-world scenarios, uniquely featuring reasoning-oriented tasks with extreme spatial sparsity (avg. area $<0.7\%$), necessitating precise visual operations and faithful visual thinking.
\textbf{2)} We propose a process-verifiable evaluation paradigm that shifts from outcome-oriented metrics to a dual-axis capability matrix, enabling fine-grained analysis to differentiate grounding failures from reasoning deficiencies.
\textbf{3)} Our extensive evaluation of state-of-the-art agentic VLMs reveals several counter-intuitive failure modes, offering actionable insights for developing more robust agentic vision models.

\section{Related Work}
\subsection{Vision-Language Models}
The landscape of Vision-Language Models (VLMs) has evolved from early alignment-based models to powerful large-scale multimodal systems. End-to-end models, such as the GPT series \citep{gpt4o}, Gemini series \cite{gemini2.0flash,gemini,gemini2.5pro} and open-source leaders like InternVL \citep{internvl3,internvl3.5}, LLaVA-OneVision \citep{llavaov} and Qwen-VL \cite{qwen2.5vl,qwen3vl}, typically process images through a fixed-resolution vision encoder. While these models have demonstrated remarkable zero-shot capabilities, they often suffer from "visual forgetting" when dealing with high-resolution images due to the information loss inherent in global downsampling \cite{hrbench,li2025multimodal}. Recent efforts have attempted to mitigate this by scaling parameters or incorporating mixture-of-experts (MoE) architectures \cite{qwen3vl}, yet the underlying black-box nature of their perception remains a significant hurdle for verifiable reasoning.

\subsection{Thinking-with-Images}
To overcome the resolution bottleneck, a new paradigm \cite{zhang2025chain,zhu2025active,li2024dtllm} known as "Thinking-with-Images" has emerged, manifesting primarily through agentic models. These systems, such as Pixel Reasoner \citep{pixelreasoner} and the Qwen3-VL series \cite{qwen3vl}, empower VLMs with the autonomy to dynamically interact with their visual input. By invoking external tools for zooming, these models can focus on task-relevant regions to capture fine-grained details that are sub-perceptual in global views \cite{minio3,deepeyes,thyme}. While this active perception mimics human saccadic eye movements and enables superior performance in dense scenes, it also introduces a new layer of complexity: the need to verify whether the model's visual operations are logically aligned with its final conclusions \cite{li2025look,hu2024fiova}.

\subsection{Thinking-with-Images Evaluation}
As the community shifts toward agentic vision, existing multimodal benchmarks \cite{yue2024mmmu,yue2025mmmu,li2025verifybench,li2025causalstep} have become increasingly inadequate for auditing the reasoning process. Traditional high-resolution benchmarks like InfoVQA \cite{infovqa} and HRBench \cite{hrbench} focus predominantly on perception-heavy tasks (e.g., OCR or object counting) without evaluating complex logical reasoning. While recent efforts like Argus Inspection \cite{yao2025argus} attempt to bridge this gap by incorporating real-world commonsense for causal reasoning, these evaluations still largely focus on the final output. Although V* Bench \cite{seal} and VisualProbe \citep{minio3} highlight the necessity of zooming, they rely solely on outcome-oriented metrics (accuracy), failing to account for cases where a model arrives at the correct answer through lucky guesses rather than precise grounding. Furthermore, to our knowledge, none of the existing suites provide expert-annotated gold BBox to verify the accuracy of visual operations in the model's "thinking" process. 
By introducing fine-grained metrics, ViEBench provides the first comprehensive framework to ensure that the "Thinking-with-Images" process is both transparent and process-verifiable.

\begin{figure*}[t!]
    \centering
    \includegraphics[width=\textwidth]{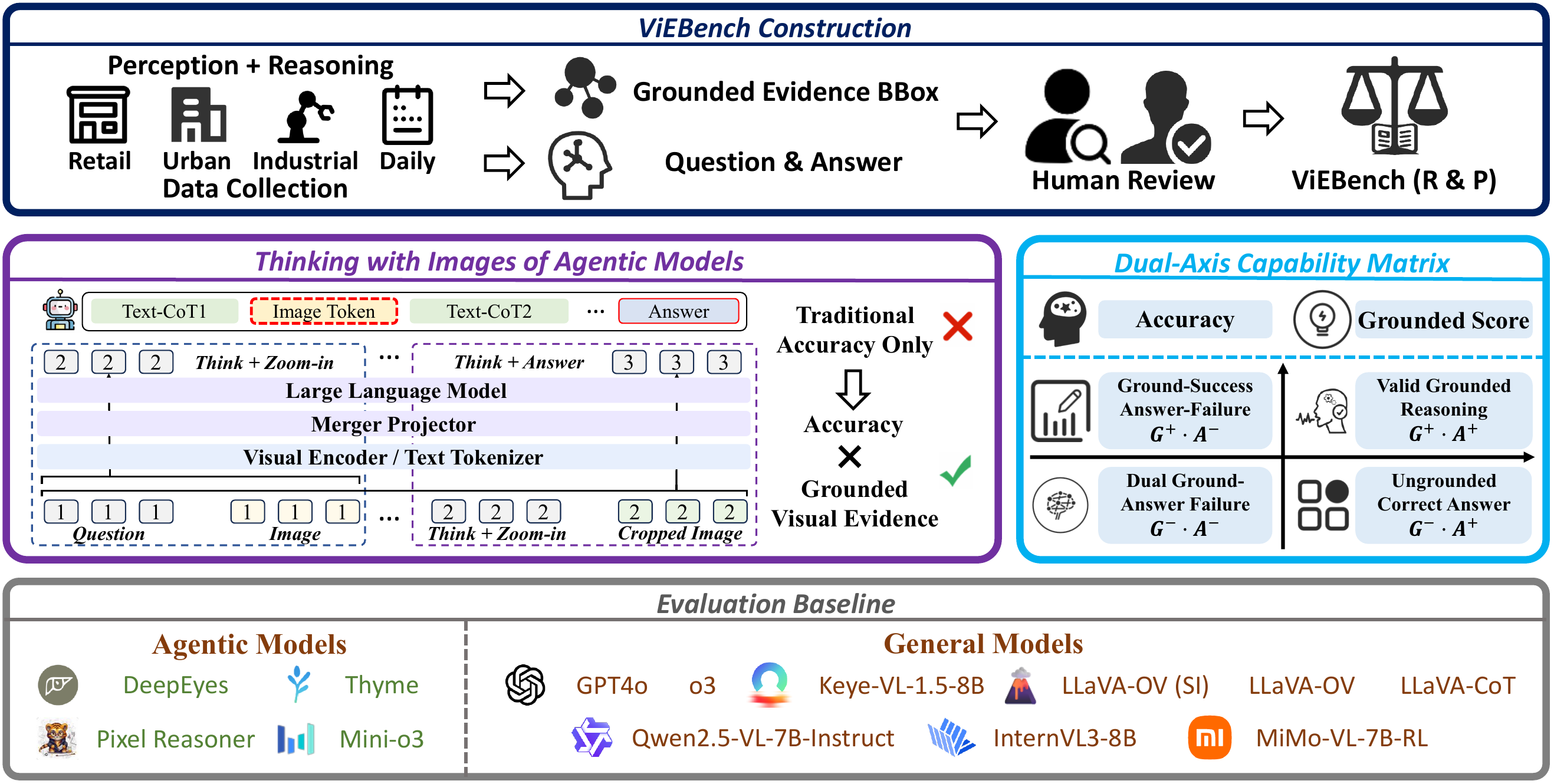}
    \vspace{-15pt}
    \caption{ViEBench audits the consistency between visual grounding and logical reasoning in agentic VLMs. By integrating expert-annotated BBox with a dual-axis evaluation protocol, we categorize model behaviors into four diagnostic metrics, providing a more rigorous assessment beyond accuracy only.}
    \label{fig:overview}
    \vspace{-15pt}
\end{figure*}

\section{ViEBench}
\subsection{Overview}

The core philosophy of ViEBench is to transition VLMs evaluation from an outcome-oriented black box to a process-verifiable diagnostic framework by isolating the interplay between visual localization and logical reasoning. This is achieved through a dual-axis capability audit that evaluates models along the orthogonal dimensions of reasoning integrity and grounding precision, allowing researchers to decouple a model's ability to locate evidence from its ability to interpret it. As shown in Fig. \ref{fig:overview}, by deconstructing performance into diagnostic quadrants, ViEBench identifies specific failure modes that standard accuracy metrics often mask. This structural approach is supported by a high-quality benchmark spanning four real-world scenarios where task-critical evidence is sub-perceptual and necessitates a verifiable "Thinking-with-Images" process.

\subsection{Dual-Axis Capability Matrix}
To systematically audit the alignment between a model's reasoning and its visual operations, we propose the dual-axis capability matrix. This framework evaluates each task instance along two orthogonal dimensions: the correct answer axis, which measures the correctness of the final textual answer, and the valid grounding axis, which quantifies the precision of the model's generated visual crops. By mapping model performance into this two-dimensional space, we define a taxonomy of four functional quadrants. This matrix allows us to move beyond holistic accuracy and specifically isolate "hallucinatory reasoning," where a model arrives at a correct conclusion despite focusing on irrelevant or misleading image regions.

We define a set of metrics based on the Intersection-over-Area (IoA) \citep{ioa} between the model's generated crop ($B_{pred}$) and the ground-truth gold BBox ($B_{gt}$). Unlike the standard Intersection-over-Union (IoU) \citep{rezatofighi2019generalized}, the IoA metric provides a more nuanced measure of spatial inclusion. Specifically,
\begin{align*}
IoA(B_{pred}, B_{gt}) = \frac{\text{Area}(B_{pred}) \cap \text{Area}(B_{gt})}{\text{Area}(B_{gt})}
\end{align*}
quantifies the extent to which the target evidence is covered by the model's crop, while the reverse formulation, 
\begin{align*}
IoA(B_{gt}, B_{pred}) = \frac{\text{Area}(B_{pred}) \cap \text{Area}(B_{gt})}{\text{Area}(B_{pred})}
\end{align*}
measures the concentration of the target within the crop area. To ensure a robust assessment that accounts for both precise tight crops and conservative expansive crops, we define the final IoA score as the maximum of these two directional metrics: 
\begin{align*}
IoA = \max(IoA(B_{pred}, B_{gt}), IoA(B_{gt}, B_{pred})).
\end{align*}
To facilitate a fine-grained analysis, we categorize the model's performance into four quadrants based on grounding ($G$) and answer ($A$) consistency: $G^+$ and $G^-$ denote successful ($IoA > 0.5$) and failed ($IoA \le 0.5$) grounding respectively, while $A^+$ and $A^-$ indicate correct and incorrect textual answers. As shown in Fig.~\ref{fig:overview}, we propose the following metrics:
\begin{itemize}[itemsep=2pt, topsep=0pt, parsep=2pt]
\item \textbf{Accuracy (Acc.):} The percentage of queries where the final textual answer is correct, regardless of the grounding quality.
\item \textbf{Grounded Score (GS):} The percentage of samples achieving successful grounding, representing the model's fundamental reliability in locating evidence.
\item \textbf{Valid Grounded Reasoning ($G^+ \cdot A^+$):} The ratio of samples where $IoA > 0.5$ \textit{and} the answer is correct. This is the primary metric for verifiable reasoning.
\item \textbf{Ground-Success Answer-Failure ($G^+ \cdot A^-$):} The ratio of samples where $IoA > 0.5$ \textit{but} the answer is incorrect.
\item \textbf{Ungrounded Correct Answer ($G^- \cdot A^+$):} The ratio of samples where $IoA \le 0.5$ \textit{but} the answer is correct, indicating a reliance on textual CoT or redundant crop.
\item \textbf{Dual Ground-Answer Failure ($G^- \cdot A^-$):} The ratio of samples where both the grounding ($IoA \le 0.5$) and the answer are incorrect.
\item \textbf{Tool Ratio (TR):} The proportion of queries where the model invokes a zooming operation.
\end{itemize}

\subsection{Data Collection}
\paragraph{Tasks.}
To systematically evaluate the agentic capabilities of VLMs, we categorize our tasks into two distinct dimensions: \textit{perception} and \textit{reasoning}. \textit{Perception tasks} focus on the model's fundamental ability to locate and identify fine-grained visual elements within high-resolution inputs. 
In contrast, \textit{reasoning tasks}—a distinctive feature of ViEBench—require that the model not only identify visual details, but also integrate these visual cues with prior knowledge and execute multi-step logical reasoning to derive correct answers.

\paragraph{Image Sources and Scenarios.}
Our data collection process is designed to capture the complexity and diversity of real-world visual reasoning tasks by curating a representative set of images sourced from both extensive web searches and the VisualProbe \citep{minio3}. These images are categorized into four distinct scenarios, including \textit{ retail, urban, industry} and \textit{daily life,} selected primarily because they represent high-stakes environments where the reliability of a model's visual grounding is paramount. 
Furthermore, these scenarios provide a rich spectrum of visual scales, ranging from the cluttered, fine-grained environments of retail shelves to the expansive, multi-object scenes of urban landscapes, thereby offering a comprehensive testbed for spatial reasoning.

\subsection{Annotation and Human Review}
The annotation process for ViEBench involves a multi-stage pipeline to ensure the highest level of ground-truth reliability. For each selected image, professional annotators are tasked with identifying the "minimal indispensable evidence" required to answer the associated query. This evidence is enclosed in a precise gold BBox, which serves as the spatial reference for our IoA-based audit. In addition to spatial grounding, annotators provide the ground-truth answer and categorize the task as either perception-heavy (requiring fine-grained identification) or reasoning-heavy (requiring multi-step logical integration). To guarantee quality, a secondary team of senior reviewers performs a verification of every instance. Any sample where the gold BBox is ambiguous or the reasoning chain is deemed non-verifiable is either refined or discarded, ensuring that every task in ViEBench presents a clear and objective challenge for the model.

\begin{figure}[t!]
    \centering
    \includegraphics[width=0.48\textwidth]{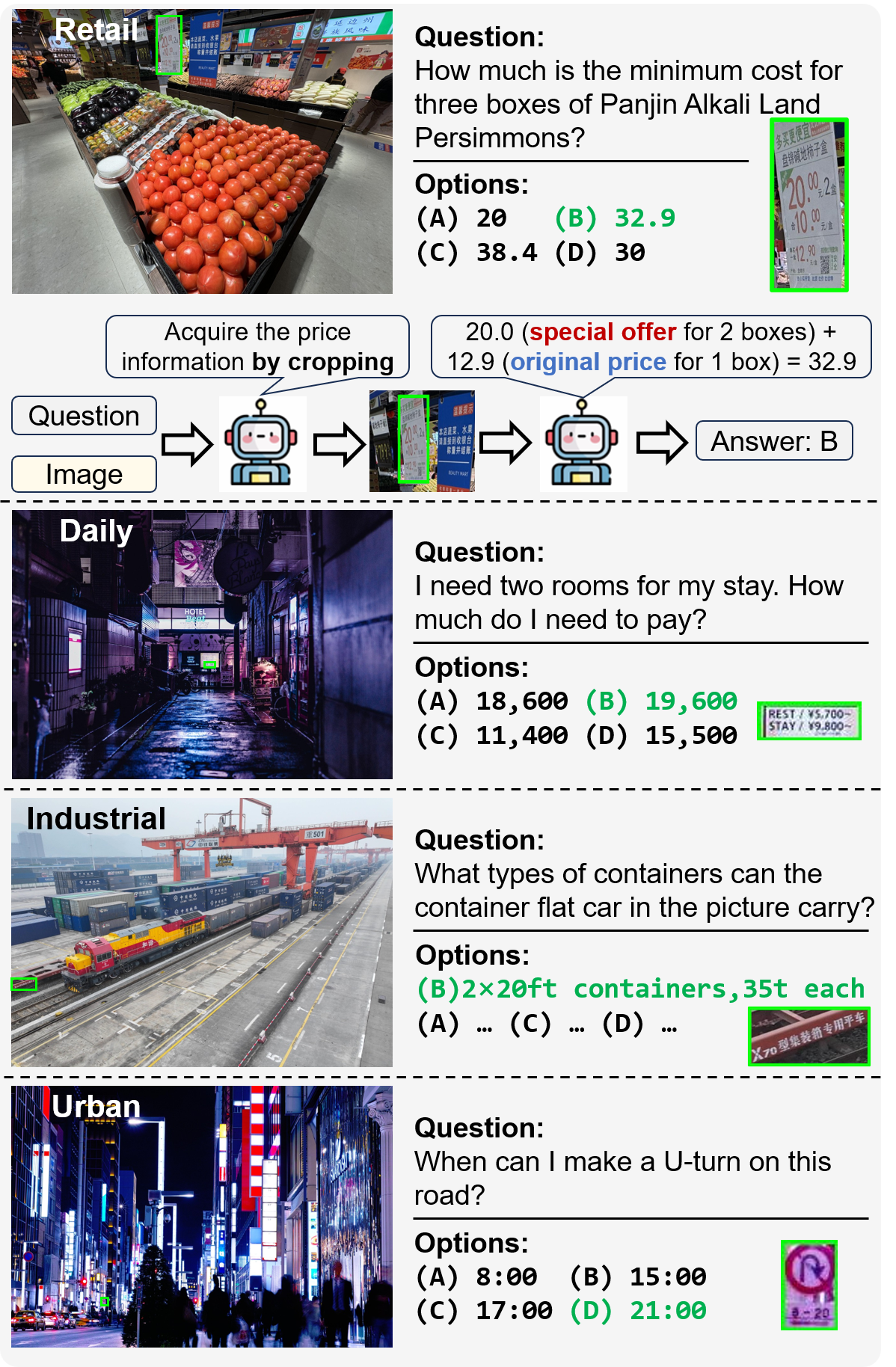}
    \vspace{-15pt}
    \caption{Representative examples from ViEBench across four real-world scenarios. Each case illustrates a complex reasoning task where the critical evidence is spatially sparse and requires precise cropping to resolve.}
    \label{fig:case}
    \vspace{-15pt}
\end{figure}

\begin{figure}[htbp!]
    \centering
    \includegraphics[width=0.35\textwidth]{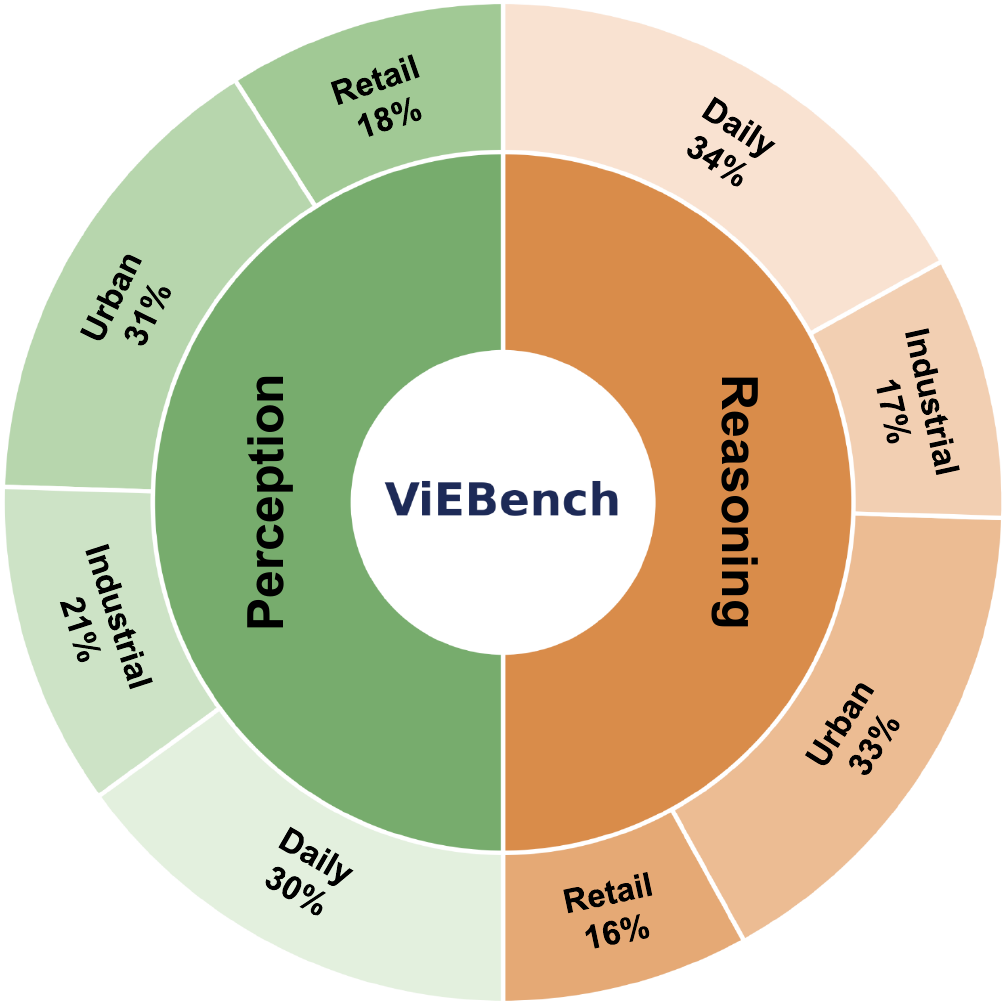}
    \vspace{-5pt}
    \caption{Scene distribution of the perception and reasoning categories in ViEBench.}
    \label{fig:statistics}
    \vspace{-15pt}
\end{figure}

\begin{table*}[htbp]
\centering
\caption{Performance of Models with Tools (Agentic Models). Inst. denotes Instruction-tuned models.}
\vspace{-5pt}
\label{tab:models_w_tools}
\resizebox{\textwidth}{!}{
\begin{tabular}{l|ccccccc}
\toprule
\textbf{Model} & 
\textbf{\makecell{Accuracy \\ $\boldsymbol{Acc.}$ $\uparrow$}} &  
\textbf{\makecell{Grounded\\Score \\ $\boldsymbol{GS}$ $\uparrow$}} & 
\textbf{\makecell{Valid Grounded\\Reasoning \\ $\boldsymbol{G^+ \cdot A^+}$ $\uparrow$}} & 
\textbf{\makecell{Ground-Success\\Answer-Failure \\ $\boldsymbol{G^+ \cdot A^-}$ $\downarrow$}} & 
\textbf{\makecell{Ungrounded\\Correct Answer \\ $\boldsymbol{G^- \cdot A^+}$ $\downarrow$}} & 
\textbf{\makecell{Dual Ground-\\Answer Failure \\ $\boldsymbol{G^- \cdot A^-}$ $\downarrow$}} & 
\textbf{\makecell{Tool\\Ratio \\ $\boldsymbol{TR}$}} \\
\midrule
\multicolumn{8}{c}{\textit{Perception}} \\
\midrule
Pixel Reasoner & 77\% & 65\% & 54\% & 12\% & 24\% & \textbf{11}\% & 95\% \\
Thyme & 77\% & 38\% & 38\% & \textbf{0}\% & 44\% & 19\% & 16\% \\
DeepEyes & 79\% & 44\% & 41\% & 3\% & 38\% & 18\% & 100\% \\
Mini-o3 & 73\% & \textbf{78\%} & 65\% & 10\% & \textbf{8\%} & 17\% & 97\% \\
Qwen3-VL-8B-Inst. & 74\% & 73\% & 66\% & 7\% & 9\% & 17\% & 98\% \\
Qwen3-VL-235B-A22B-Inst. & 73\% & 69\% & 62\% & 7\% & 11\% & 20\% & 100\% \\
Qwen3-VL-32B-Inst. & \textbf{81\%} & 75\% & \textbf{71\%} & 4\% & 9\% & 16\% & 93\% \\
\midrule
\multicolumn{8}{c}{\textit{Reasoning}} \\
\midrule
Pixel Reasoner & 59\% & 64\% & 40\% & 23\% & 23\% & 15\% & 80\% \\
Thyme & 69\% & 29\% & 29\% & \textbf{0\%} & 57\% & \textbf{14\%} & 7\% \\
DeepEyes & 60\% & 40\% & 27\% & 13\% & 33\% & 27\% & 100\% \\
Mini-o3 & 58\% & \textbf{78\%} & 47\% & 28\% & \textbf{11\%} & \textbf{14\%} & 97\% \\
Qwen3-VL-8B-Inst. & 71\% & 67\% & \textbf{57\%} & 10\% & 15\% & 18\% & 92\% \\
Qwen3-VL-235B-A22B-Inst. & 71\% & 66\% & 54\% & 13\% & 18\% & 16\% & 95\% \\
Qwen3-VL-32B-Inst. & \textbf{74\%} & 68\% & 56\% & 13\% & 17\% & 15\% & 95\% \\
\bottomrule
\end{tabular}
}
\vspace{-10pt}
\end{table*}

\subsection{Statistics}
The finalized ViEBench benchmark consists of 200 high-resolution multiple-choice QA pairs, meticulously curated to ensure a balanced distribution across scenarios and cognitive demands. Quantitatively, the dataset is perfectly bifurcated into perception (50\%) and reasoning (50\%) tasks, providing an even ground for evaluating both fine-grained recognition and complex logical reasoning across four key real-world scenarios: urban (32\%), daily life (32\%), industrial (19\%) and retail (17\%). We provide some representative examples in Fig. \ref{fig:case}.

A defining characteristic of ViEBench is the extreme spatial sparsity of task-critical evidence, a design choice specifically intended to necessitate active "Thinking-with-Images" behaviors. The expert-annotated gold BBox occupies a very small proportion of the total image area, averaging only 0.32\% for perception-based queries and 0.63\% for reasoning-based ones. This deliberate concentration of information ensures that essential visual cues remain sub-perceptual in standard global downsamplings, thereby compelling models to execute precise local zooming and cropping to resolve the evidence. To ensure the integrity of the benchmark, every instance was produced through a rigorous pipeline involving exhaustive expert manual annotation, resulting in a diagnostic suite that is both empirically challenging and process-verifiable.

\section{Experiment}
\subsection{Baselines}
To provide a comprehensive benchmark of current VLM capabilities, we evaluate two distinct categories of models:

\textbf{Models with Tools (Agentic Models).} This category comprises agentic systems that utilize external tools for zooming operations. These include Pixel Reasoner \citep{pixelreasoner}, Thyme \citep{thyme}, DeepEyes \citep{deepeyes}, Mini-o3 \citep{minio3}, Qwen3-VL-8B-Instruct, Qwen3-VL-235B-A22B-Instruct and Qwen3-VL-32B-Instruct \cite{qwen3vl}. These models are evaluated on their ability to strategically invoke tools to locate evidence before generating a final answer.

\textbf{Models without Tools (End-to-end VLMs).} This category includes state-of-the-art general-purpose VLMs and models specifically optimized for CoT reasoning \citep{cot}. We evaluate proprietary frontiers such as GPT-4o \citep{gpt4o} and o3 \cite{o3}, alongside leading open-source models including Qwen2.5-VL-7B-Instruct \citep{qwen2.5vl}, InternVL3-8B \citep{internvl3}, and LLaVA-OneVision (OV) \citep{llavaov}. Specifically, for the LLaVA series, we include both the standard LLaVA-OV and its LLaVA-OV (SI) variant. We also incorporate specialized reasoning models such as LLaVA-CoT \citep{llavacot}, Keye-VL-1.5-8B \citep{keyevl}, and MiMo-VL-7B-RL \citep{mimovl}, which are designed to enhance the depth of reasoning.

\subsection{Evaluation Protocol}
We employ distinct evaluation pipelines and reporting scopes for the two categories of baselines. For models with tools (agentic models), we strictly adhere to the evaluation settings and environment configurations specified in their respective official repositories to ensure tools are invoked as intended; for these models, we report the full set of seven metrics to perform a comprehensive process-level audit. In contrast, for models without tools (end-to-end VLMs), we utilize the VLMEvalKit \citep{vlmevalkit} framework to ensure a fair comparison. Since these end-to-end models lack an explicit cropping mechanism to expose their internal focus, we report only their overall accuracy.

\begin{table}[t]
\centering
\caption{Performance of Models without Tools (End-to-end VLMs). (Inst. denotes Instruction-tuned)}
\vspace{-5pt}
\label{tab:models_wo_tools}
\resizebox{0.45\textwidth}{!}{%
\begin{tabular}{lcc}
\toprule
\multirow{2}{*}{\textbf{Model}} & \multicolumn{2}{c}{\textbf{Accuracy}} \\
\cmidrule(lr){2-3}
 & \textbf{Perception} & \textbf{Reasoning} \\
\midrule
GPT4o & 66\% & 64\% \\
o3 & 71\% & 69\% \\
Qwen2.5-VL-7B-Inst. & 74\% & 58\% \\
InternVL3 & 75\% & 63\% \\
LLaVA-CoT & 51\% & 49\% \\
LLaVA-OV (SI) & 62\% & 63\% \\
LLaVA-OV & 62\% & 56\% \\
Keye-VL-1.5-8B & 72\% & 65\% \\
MiMo-VL-7B-RL & 71\% & 60\% \\
\bottomrule
\end{tabular}%
}
\vspace{-10pt}
\end{table}

\begin{figure*}[t!]
    \centering
    \includegraphics[width=0.85\textwidth]{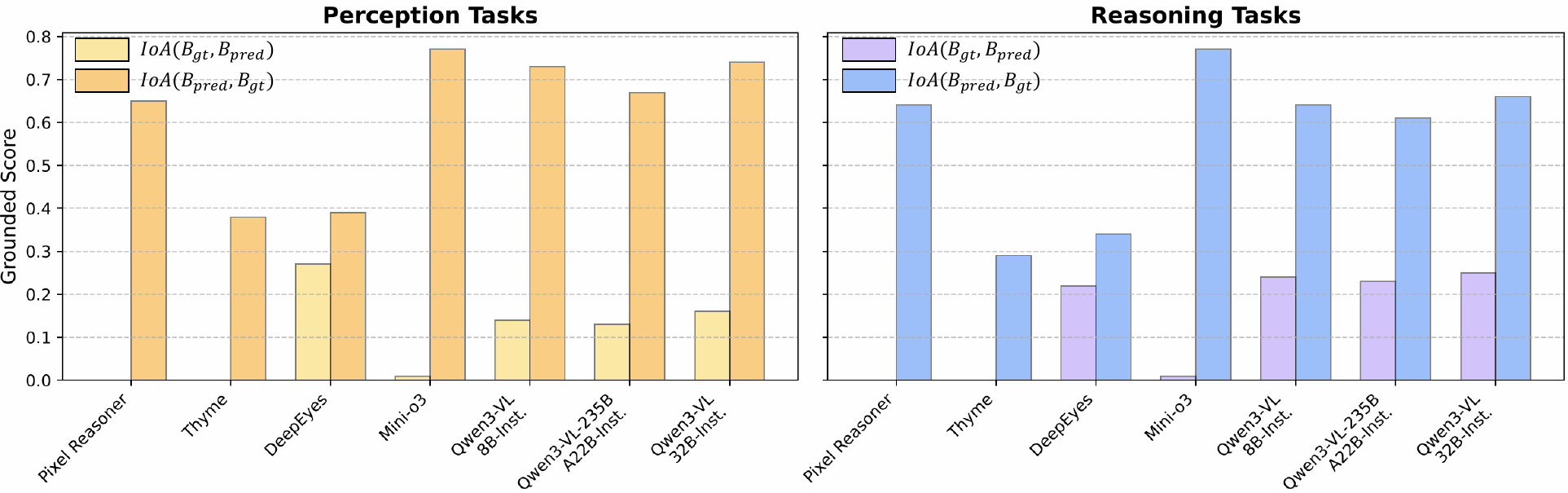}
    \caption{We visualize $IoA(B_{gt}, B_{pred})$ and $IoA(B_{pred}, B_{gt})$ across perception and reasoning tasks. The results reveal distinct strategies. (Inst. denotes Instruction-tuned)}
    \label{fig:bar_chart}
    \vspace{-10pt}
\end{figure*}

\section{Main Results and Analysis}
Tab. \ref{tab:models_w_tools} and Tab. \ref{tab:models_wo_tools} provide a comprehensive evaluation of state-of-the-art VLMs. Due to the restricted accessibility of internal cropping results from proprietary closed-source models, our analysis primarily focuses on open-source agentic models; however, developers and model hosts can readily apply this auditing framework to diagnose and refine the reasoning behaviors of any specific model.

\subsection{Necessity of Reasoning-centric Evaluation}
The comparative results between perception and reasoning tasks in Tab. \ref{tab:models_w_tools} demonstrate the necessity of ViEBench, as conventional benchmarks measuring only final accuracy fail to capture the capability collapse that occurs when task complexity increases. On simpler perception-oriented tasks, the performance gap between models is relatively narrow, and accuracy remains high; for instance, Mini-o3 \cite{minio3} achieves 73\%. However, on complex reasoning queries, its Accuracy (Acc.) drops significantly to 58\%. Crucially, this decline occurs despite the model maintaining an identical Grounded Score (GS) of 78\% across both categories. This decoupling of localization success and final correctness suggests that while the model's perception remains stable, the reasoning demands of ViEBench expose a latent reasoning bottleneck. Without a specialized reasoning-centric benchmark, such a significant drop in performance would be hidden within total accuracy scores, making ViEBench essential for identifying the limits of agentic models beyond basic recognition.

\subsection{Fine-grained Audit via Capability Matrix}

By deconstructing performance into four $G/A$ quadrants, ViEBench enables a transparent audit of reasoning processes indistinguishable through accuracy alone. The Semantic Reasoning Bottleneck characterizes models with superior perception but deficient logic; for instance, in reasoning tasks, Mini-o3 \cite{minio3} achieves the highest GS (78\%) yet exhibits a peak Ground-Success and Answer-Failure ($G^+ \cdot A^-$) rate (28\%). This proves failure stems not from blind search, but from an inability to synthesize localized cues into correct conclusions. Conversely, Superficial Correctness exposes models like DeepEyes \cite{deepeyes} with high Ungrounded Correct Answer ($G^- \cdot A^+$) rates, suggesting significant redundant cropping where correct answers are reached despite misplaced visual focus. Finally, Grounded Reasoning Integrity confirms the reliability of top-tier models like Qwen3-VL-32B-Instruct \cite{qwen3vl}, which achieves a high Valid Grounded Reasoning ($G^+ \cdot A^+$) score (71\% in perception) with minimal $G^- \cdot A^+$ (9\%), proving its success derives from faithful visual evidence. This mapping precisely identifies whether a model’s bottleneck lies in perception or reasoning integration. This diagnostic depth enables ViEBench to expose failure modes that remain invisible to traditional benchmarks.

\subsection{Adaptive Thinking and Tool Efficiency}

A significant finding is the variation in adaptivity across models, particularly regarding the trade-off between tool efficiency and grounding precision. Thyme \cite{thyme} achieves a competitive Acc. (77\% in perception) while maintaining an exceptionally low Tool Ratio (TR) of 16\%, suggesting an efficient adaptive mechanism that invokes "Thinking-with-Images" only for highly ambiguous samples. However, in reasoning categories, Thyme's GS remains limited, and its $G^- \cdot A^+$ rate reaches 57\%, indicating that its tool calls frequently result in redundant cropping. For such efficient models, the path to improvement lies in recalibrating tool-invocation triggers to ensure that limited crops are precisely aligned with task-critical evidence. By reducing these redundant or misaligned operations, models can further minimize per-instance inference time and avoid potential interference from irrelevant visual noise, ultimately strengthening the $G^+ \cdot A^+$ path without sacrificing their computational advantage.

\subsection{Grounding and Reasoning Alignment}
We observe a core challenge in cognitive consistency, defined as the model's ability to maintain reasoning integrity once the correct evidence is localized. By comparing the GS and $G^+ \cdot A^+$, we find that Pixel Reasoner \cite{pixelreasoner} exhibits robust consistency in perception tasks. In these cases, it achieves a GS of 65\%, and a high proportion of these samples (54\%) are successfully converted into $G^+ \cdot A^+$. However, its performance decays significantly in reasoning tasks, where a similar GS of 64\% only yields a $G^+ \cdot A^+$ score of 40\%. This decay indicates that while the model’s spatial search capability is consistent across task types, its reasoning capability remains fragile when integrating visual evidence. Localization is a necessary but insufficient condition for reasoning; the substantial gap observed in reasoning tasks suggests that the model often identifies the correct evidence but fails to construct a reliable CoT for complex queries.

\subsection{Spatial Alignment and Crop Strategies}

The bidirectional IoA analysis in Fig. \ref{fig:bar_chart} provides deeper insights into the specific "Thinking-with-Images" behaviors of different models, revealing that a tighter crop does not inherently guarantee superior reasoning. Models such as Mini-o3 \cite{minio3} and the Qwen3-VL series \cite{qwen3vl} generally exhibit an expansive coverage strategy, characterized by high $IoA(B_{pred}, B_{gt})$ values results. This pattern indicates that while their generated crops are relatively large compared to the target evidence, they successfully encompass the entire gold BBox. Crucially, the Qwen3-VL series demonstrates that such moderate spatial redundancy is not a hindrance; by effectively utilizing visual cues within these larger crops, Qwen3-VL-32B-Instruct achieves high $G^+ \cdot A^+$ and low $G^+ \cdot A^-$ rates. Conversely, models like DeepEyes \cite{deepeyes} often generate more concentrated crops with higher $IoA(B_{gt}, B_{pred})$ levels, yet this tighter spatial focus does not translate into performance gains in reasoning accuracy. This divergence validates the design of our evaluation metric, which prioritizes the coverage of essential evidence over mere boundary precision. It further suggests that for agentic VLMs, over-optimizing for tight BBox coordinates during training may be counterproductive. Instead, the focus should remain on developing adaptive cropping mechanisms that allow models to determine an optimal viewing scale based on their reasoning capacity, ensuring that sufficient context is preserved to support the subsequent CoT.

\section{Conclusion}
In this paper, we address a critical gap in the evaluation of agentic VLMs by moving beyond simplistic outcome-oriented metrics toward a process-verifiable paradigm. Through ViEBench, we provide a rigorous diagnostic framework that leverages fine-grained perception and complex reasoning tasks to evaluate the capabilities of agentic VLMs within high-resolution environments. Our dual-axis capability matrix uniquely decomposes performance into grounding accuracy and reasoning logic, revealing that current models frequently rely on "ungrounded correct answers" or struggle to synthesize evidence even after successful localization. These findings underscore that the next frontier for VLMs lies in achieving cognitive consistency between visual perception and logical inference. 

\clearpage
\section*{Limitations}
While ViEBench provides a rigorous and process-verifiable framework for auditing agentic VLMs, our current benchmark primarily focuses on cropping as the central visual operation for "Thinking-with-Images," as it represents the most critical mechanism for resolving spatial sparsity and fine-grained perception. However, as the field evolves, agentic models are expected to perform a broader suite of complex visual operations. Since ViEBench is currently optimized for high-resolution spatial perception and reasoning tasks, it does not fully account for the evaluation of these emerging diverse tool-use capabilities. We recognize this as a vital area for growth and plan to incorporate the assessment of more varied visual operations into our subsequent work to maintain a comprehensive diagnostic standard for future multimodal agents.

\bibliography{custom}

@article{minio3,
  title={Mini-o3: Scaling up reasoning patterns and interaction turns for visual search},
  author={Lai, Xin and Li, Junyi and Li, Wei and Liu, Tao and Li, Tianjian and Zhao, Hengshuang},
  journal={arXiv preprint arXiv:2509.07969},
  year={2025}
}

@inproceedings{ioa,
  title={Objectseeker: Certifiably robust object detection against patch hiding attacks via patch-agnostic masking},
  author={Xiang, Chong and Valtchanov, Alexander and Mahloujifar, Saeed and Mittal, Prateek},
  booktitle={2023 IEEE Symposium on Security and Privacy (SP)},
  pages={1329--1347},
  year={2023},
  organization={IEEE}
}

@article{cot,
  title={Chain-of-thought prompting elicits reasoning in large language models},
  author={Wei, Jason and Wang, Xuezhi and Schuurmans, Dale and Bosma, Maarten and Xia, Fei and Chi, Ed and Le, Quoc V and Zhou, Denny and others},
  journal={Advances in neural information processing systems},
  volume={35},
  pages={24824--24837},
  year={2022}
}

@article{gpt4o,
  title={Gpt-4o system card},
  author={Hurst, Aaron and Lerer, Adam and Goucher, Adam P and Perelman, Adam and Ramesh, Aditya and Clark, Aidan and Ostrow, AJ and Welihinda, Akila and Hayes, Alan and Radford, Alec and others},
  journal={arXiv preprint arXiv:2410.21276},
  year={2024}
}

@article{qwen2.5vl,
  title={Qwen2. 5-vl technical report},
  author={Bai, Shuai and Chen, Keqin and Liu, Xuejing and Wang, Jialin and Ge, Wenbin and Song, Sibo and Dang, Kai and Wang, Peng and Wang, Shijie and Tang, Jun and others},
  journal={arXiv preprint arXiv:2502.13923},
  year={2025}
}

@article{internvl3,
  title={Internvl3: Exploring advanced training and test-time recipes for open-source multimodal models},
  author={Zhu, Jinguo and Wang, Weiyun and Chen, Zhe and Liu, Zhaoyang and Ye, Shenglong and Gu, Lixin and Tian, Hao and Duan, Yuchen and Su, Weijie and Shao, Jie and others},
  journal={arXiv preprint arXiv:2504.10479},
  year={2025}
}

@article{llavaov,
  title={Llava-onevision: Easy visual task transfer},
  author={Li, Bo and Zhang, Yuanhan and Guo, Dong and Zhang, Renrui and Li, Feng and Zhang, Hao and Zhang, Kaichen and Zhang, Peiyuan and Li, Yanwei and Liu, Ziwei and others},
  journal={arXiv preprint arXiv:2408.03326},
  year={2024}
}

@inproceedings{llavacot,
  title={Llava-cot: Let vision language models reason step-by-step},
  author={Xu, Guowei and Jin, Peng and Wu, Ziang and Li, Hao and Song, Yibing and Sun, Lichao and Yuan, Li},
  booktitle={Proceedings of the IEEE/CVF International Conference on Computer Vision},
  pages={2087--2098},
  year={2025}
}

@article{keyevl,
  title={Kwai keye-vl 1.5 technical report},
  author={Yang, Biao and Wen, Bin and Ding, Boyang and Liu, Changyi and Chu, Chenglong and Song, Chengru and Rao, Chongling and Yi, Chuan and Li, Da and Zang, Dunju and others},
  journal={arXiv preprint arXiv:2509.01563},
  year={2025}
}

@article{mimovl,
  title={MiMo: Unlocking the Reasoning Potential of Language Model--From Pretraining to Posttraining},
  author={Xiaomi, LLM and Xia, Bingquan and Shen, Bowen and Zhu, Dawei and Zhang, Di and Wang, Gang and Zhang, Hailin and Liu, Huaqiu and Xiao, Jiebao and Dong, Jinhao and others},
  journal={arXiv preprint arXiv:2505.07608},
  year={2025}
}

@article{pixelreasoner,
  title={Pixel reasoner: Incentivizing pixel-space reasoning with curiosity-driven reinforcement learning},
  author={Su, Alex and Wang, Haozhe and Ren, Weiming and Lin, Fangzhen and Chen, Wenhu},
  journal={arXiv preprint arXiv:2505.15966},
  year={2025}
}

@article{thyme,
  title={Thyme: Think Beyond Images},
  author={Zhang, Yi-Fan and Lu, Xingyu and Yin, Shukang and Fu, Chaoyou and Chen, Wei and Hu, Xiao and Wen, Bin and Jiang, Kaiyu and Liu, Changyi and Zhang, Tianke and others},
  journal={arXiv preprint arXiv:2508.11630},
  year={2025}
}

@article{deepeyes,
  title={DeepEyes: Incentivizing" Thinking with Images" via Reinforcement Learning},
  author={Zheng, Ziwei and Yang, Michael and Hong, Jack and Zhao, Chenxiao and Xu, Guohai and Yang, Le and Shen, Chao and Yu, Xing},
  journal={arXiv preprint arXiv:2505.14362},
  year={2025}
}

@misc{qwen3vl,
      title={Qwen3-VL Technical Report}, 
      author={Shuai Bai and Yuxuan Cai and Ruizhe Chen and Keqin Chen and Xionghui Chen and Zesen Cheng and Lianghao Deng and Wei Ding and Chang Gao and Chunjiang Ge and Wenbin Ge and Zhifang Guo and Qidong Huang and Jie Huang and Fei Huang and Binyuan Hui and Shutong Jiang and Zhaohai Li and Mingsheng Li and Mei Li and Kaixin Li and Zicheng Lin and Junyang Lin and Xuejing Liu and Jiawei Liu and Chenglong Liu and Yang Liu and Dayiheng Liu and Shixuan Liu and Dunjie Lu and Ruilin Luo and Chenxu Lv and Rui Men and Lingchen Meng and Xuancheng Ren and Xingzhang Ren and Sibo Song and Yuchong Sun and Jun Tang and Jianhong Tu and Jianqiang Wan and Peng Wang and Pengfei Wang and Qiuyue Wang and Yuxuan Wang and Tianbao Xie and Yiheng Xu and Haiyang Xu and Jin Xu and Zhibo Yang and Mingkun Yang and Jianxin Yang and An Yang and Bowen Yu and Fei Zhang and Hang Zhang and Xi Zhang and Bo Zheng and Humen Zhong and Jingren Zhou and Fan Zhou and Jing Zhou and Yuanzhi Zhu and Ke Zhu},
      year={2025},
      eprint={2511.21631},
      archivePrefix={arXiv},
      primaryClass={cs.CV},
      url={https://arxiv.org/abs/2511.21631}, 
}

@inproceedings{vlmevalkit,
  title={Vlmevalkit: An open-source toolkit for evaluating large multi-modality models},
  author={Duan, Haodong and Yang, Junming and Qiao, Yuxuan and Fang, Xinyu and Chen, Lin and Liu, Yuan and Dong, Xiaoyi and Zang, Yuhang and Zhang, Pan and Wang, Jiaqi and others},
  booktitle={Proceedings of the 32nd ACM International Conference on Multimedia},
  pages={11198--11201},
  year={2024}
}

@article{gemini2.0flash,
  title={Gemini 1.5: Unlocking multimodal understanding across millions of tokens of context},
  author={Team, Gemini and Georgiev, Petko and Lei, Ving Ian and Burnell, Ryan and Bai, Libin and Gulati, Anmol and Tanzer, Garrett and Vincent, Damien and Pan, Zhufeng and Wang, Shibo and others},
  journal={arXiv preprint arXiv:2403.05530},
  year={2024}
}

@article{gemini2.5pro,
  title={Gemini 2.5: Pushing the frontier with advanced reasoning, multimodality, long context, and next generation agentic capabilities},
  author={Comanici, Gheorghe and Bieber, Eric and Schaekermann, Mike and Pasupat, Ice and Sachdeva, Noveen and Dhillon, Inderjit and Blistein, Marcel and Ram, Ori and Zhang, Dan and Rosen, Evan and others},
  journal={arXiv preprint arXiv:2507.06261},
  year={2025}
}

@article{gemini,
  title={Gemini: a family of highly capable multimodal models},
  author={Team, Gemini and Anil, Rohan and Borgeaud, Sebastian and Alayrac, Jean-Baptiste and Yu, Jiahui and Soricut, Radu and Schalkwyk, Johan and Dai, Andrew M and Hauth, Anja and Millican, Katie and others},
  journal={arXiv preprint arXiv:2312.11805},
  year={2023}
}

@article{internvl3.5,
  title={InternVL3. 5: Advancing Open-Source Multimodal Models in Versatility, Reasoning, and Efficiency},
  author={Wang, Weiyun and Gao, Zhangwei and Gu, Lixin and Pu, Hengjun and Cui, Long and Wei, Xingguang and Liu, Zhaoyang and Jing, Linglin and Ye, Shenglong and Shao, Jie and others},
  journal={arXiv preprint arXiv:2508.18265},
  year={2025}
}

@inproceedings{seal,
  title={V?: Guided visual search as a core mechanism in multimodal llms},
  author={Wu, Penghao and Xie, Saining},
  booktitle={Proceedings of the IEEE/CVF Conference on Computer Vision and Pattern Recognition},
  pages={13084--13094},
  year={2024}
}

@article{hrbench,
      title={Divide, Conquer and Combine: A Training-Free Framework for High-Resolution Image Perception in Multimodal Large Language Models}, 
      author={Wenbin Wang and Liang Ding and Minyan Zeng and Xiabin Zhou and Li Shen and Yong Luo and Dacheng Tao},
      year={2024},
      journal={arXiv preprint}
}

@inproceedings{infovqa,
  title={Infographicvqa},
  author={Mathew, Minesh and Bagal, Viraj and Tito, Rub{\`e}n and Karatzas, Dimosthenis and Valveny, Ernest and Jawahar, CV},
  booktitle={Proceedings of the IEEE/CVF Winter Conference on Applications of Computer Vision},
  pages={1697--1706},
  year={2022}
}

@article{li2025look,
  title={Look Less, Reason More: Rollout-Guided Adaptive Pixel-Space Reasoning},
  author={Li, Xuchen and Li, Xuzhao and Gao, Jiahui and Pi, Renjie and Hu, Shiyu and Zhang, Wentao},
  journal={arXiv preprint arXiv:2510.01681},
  year={2025}
}

@article{su2025thinking,
  title={Thinking with images for multimodal reasoning: Foundations, methods, and future frontiers},
  author={Su, Zhaochen and Xia, Peng and Guo, Hangyu and Liu, Zhenhua and Ma, Yan and Qu, Xiaoye and Liu, Jiaqi and Li, Yanshu and Zeng, Kaide and Yang, Zhengyuan and others},
  journal={arXiv preprint arXiv:2506.23918},
  year={2025}
}

@online{o3,
  author  = {OpenAI},
  title   = {Introducing o3 and o4-mini},
  year    = {2025},
  url     = {https://openai.com/index/introducing-o3-and-o4-mini/},
  urldate = {2025-04-16}
}

@misc{kimi-k1.5,
      title={Kimi k1.5: Scaling Reinforcement Learning with LLMs}, 
      author={Kimi Team and Angang Du and Bofei Gao and Bowei Xing and Changjiu Jiang and Cheng Chen and Cheng Li and Chenjun Xiao and Chenzhuang Du and Chonghua Liao and Chuning Tang and Congcong Wang and Dehao Zhang and Enming Yuan and Enzhe Lu and Fengxiang Tang and Flood Sung and Guangda Wei and Guokun Lai and Haiqing Guo and Han Zhu and Hao Ding and Hao Hu and Hao Yang and Hao Zhang and Haotian Yao and Haotian Zhao and Haoyu Lu and Haoze Li and Haozhen Yu and Hongcheng Gao and Huabin Zheng and Huan Yuan and Jia Chen and Jianhang Guo and Jianlin Su and Jianzhou Wang and Jie Zhao and Jin Zhang and Jingyuan Liu and Junjie Yan and Junyan Wu and Lidong Shi and Ling Ye and Longhui Yu and Mengnan Dong and Neo Zhang and Ningchen Ma and Qiwei Pan and Qucheng Gong and Shaowei Liu and Shengling Ma and Shupeng Wei and Sihan Cao and Siying Huang and Tao Jiang and Weihao Gao and Weimin Xiong and Weiran He and Weixiao Huang and Weixin Xu and Wenhao Wu and Wenyang He and Xianghui Wei and Xianqing Jia and Xingzhe Wu and Xinran Xu and Xinxing Zu and Xinyu Zhou and Xuehai Pan and Y. Charles and Yang Li and Yangyang Hu and Yangyang Liu and Yanru Chen and Yejie Wang and Yibo Liu and Yidao Qin and Yifeng Liu and Ying Yang and Yiping Bao and Yulun Du and Yuxin Wu and Yuzhi Wang and Zaida Zhou and Zhaoji Wang and Zhaowei Li and Zhen Zhu and Zheng Zhang and Zhexu Wang and Zhilin Yang and Zhiqi Huang and Zihao Huang and Ziyao Xu and Zonghan Yang and Zongyu Lin},
      year={2025},
      eprint={2501.12599},
      archivePrefix={arXiv},
      primaryClass={cs.AI},
      url={https://arxiv.org/abs/2501.12599}, 
}

@misc{kimi-k2,
      title={Kimi K2: Open Agentic Intelligence}, 
      author={Kimi Team and Yifan Bai and Yiping Bao and Guanduo Chen and Jiahao Chen and Ningxin Chen and Ruijue Chen and Yanru Chen and Yuankun Chen and Yutian Chen and Zhuofu Chen and Jialei Cui and Hao Ding and Mengnan Dong and Angang Du and Chenzhuang Du and Dikang Du and Yulun Du and Yu Fan and Yichen Feng and Kelin Fu and Bofei Gao and Hongcheng Gao and Peizhong Gao and Tong Gao and Xinran Gu and Longyu Guan and Haiqing Guo and Jianhang Guo and Hao Hu and Xiaoru Hao and Tianhong He and Weiran He and Wenyang He and Chao Hong and Yangyang Hu and Zhenxing Hu and Weixiao Huang and Zhiqi Huang and Zihao Huang and Tao Jiang and Zhejun Jiang and Xinyi Jin and Yongsheng Kang and Guokun Lai and Cheng Li and Fang Li and Haoyang Li and Ming Li and Wentao Li and Yanhao Li and Yiwei Li and Zhaowei Li and Zheming Li and Hongzhan Lin and Xiaohan Lin and Zongyu Lin and Chengyin Liu and Chenyu Liu and Hongzhang Liu and Jingyuan Liu and Junqi Liu and Liang Liu and Shaowei Liu and T. Y. Liu and Tianwei Liu and Weizhou Liu and Yangyang Liu and Yibo Liu and Yiping Liu and Yue Liu and Zhengying Liu and Enzhe Lu and Lijun Lu and Shengling Ma and Xinyu Ma and Yingwei Ma and Shaoguang Mao and Jie Mei and Xin Men and Yibo Miao and Siyuan Pan and Yebo Peng and Ruoyu Qin and Bowen Qu and Zeyu Shang and Lidong Shi and Shengyuan Shi and Feifan Song and Jianlin Su and Zhengyuan Su and Xinjie Sun and Flood Sung and Heyi Tang and Jiawen Tao and Qifeng Teng and Chensi Wang and Dinglu Wang and Feng Wang and Haiming Wang and Jianzhou Wang and Jiaxing Wang and Jinhong Wang and Shengjie Wang and Shuyi Wang and Yao Wang and Yejie Wang and Yiqin Wang and Yuxin Wang and Yuzhi Wang and Zhaoji Wang and Zhengtao Wang and Zhexu Wang and Chu Wei and Qianqian Wei and Wenhao Wu and Xingzhe Wu and Yuxin Wu and Chenjun Xiao and Xiaotong Xie and Weimin Xiong and Boyu Xu and Jing Xu and Jinjing Xu and L. H. Xu and Lin Xu and Suting Xu and Weixin Xu and Xinran Xu and Yangchuan Xu and Ziyao Xu and Junjie Yan and Yuzi Yan and Xiaofei Yang and Ying Yang and Zhen Yang and Zhilin Yang and Zonghan Yang and Haotian Yao and Xingcheng Yao and Wenjie Ye and Zhuorui Ye and Bohong Yin and Longhui Yu and Enming Yuan and Hongbang Yuan and Mengjie Yuan and Haobing Zhan and Dehao Zhang and Hao Zhang and Wanlu Zhang and Xiaobin Zhang and Yangkun Zhang and Yizhi Zhang and Yongting Zhang and Yu Zhang and Yutao Zhang and Yutong Zhang and Zheng Zhang and Haotian Zhao and Yikai Zhao and Huabin Zheng and Shaojie Zheng and Jianren Zhou and Xinyu Zhou and Zaida Zhou and Zhen Zhu and Weiyu Zhuang and Xinxing Zu},
      year={2025},
      eprint={2507.20534},
      archivePrefix={arXiv},
      primaryClass={cs.LG},
      url={https://arxiv.org/abs/2507.20534}, 
}

@misc{minimax-m1,
      title={MiniMax-M1: Scaling Test-Time Compute Efficiently with Lightning Attention}, 
      author={MiniMax},
      year={2025},
      eprint={2506.13585},
      archivePrefix={arXiv},
      primaryClass={cs.CL},
      url={https://arxiv.org/abs/2506.13585}, 
}

@article{chen2024we,
  title={Are we on the right way for evaluating large vision-language models?},
  author={Chen, Lin and Li, Jinsong and Dong, Xiaoyi and Zhang, Pan and Zang, Yuhang and Chen, Zehui and Duan, Haodong and Wang, Jiaqi and Qiao, Yu and Lin, Dahua and others},
  journal={Advances in Neural Information Processing Systems},
  volume={37},
  pages={27056--27087},
  year={2024}
}

@article{zhang2025chain,
  title={Chain-of-Focus: Adaptive Visual Search and Zooming for Multimodal Reasoning via RL},
  author={Zhang, Xintong and Gao, Zhi and Zhang, Bofei and Li, Pengxiang and Zhang, Xiaowen and Liu, Yang and Yuan, Tao and Wu, Yuwei and Jia, Yunde and Zhu, Song-Chun and others},
  journal={arXiv preprint arXiv:2505.15436},
  year={2025}
}

@article{zhu2025active,
  title={Active-O3: Empowering Multimodal Large Language Models with Active Perception via GRPO},
  author={Zhu, Muzhi and Zhong, Hao and Zhao, Canyu and Du, Zongze and Huang, Zheng and Liu, Mingyu and Chen, Hao and Zou, Cheng and Chen, Jingdong and Yang, Ming and others},
  journal={arXiv preprint arXiv:2505.21457},
  year={2025}
}

@inproceedings{yue2024mmmu,
  title={Mmmu: A massive multi-discipline multimodal understanding and reasoning benchmark for expert agi},
  author={Yue, Xiang and Ni, Yuansheng and Zhang, Kai and Zheng, Tianyu and Liu, Ruoqi and Zhang, Ge and Stevens, Samuel and Jiang, Dongfu and Ren, Weiming and Sun, Yuxuan and others},
  booktitle={Proceedings of the IEEE/CVF Conference on Computer Vision and Pattern Recognition},
  pages={9556--9567},
  year={2024}
}

@inproceedings{yue2025mmmu,
  title={Mmmu-pro: A more robust multi-discipline multimodal understanding benchmark},
  author={Yue, Xiang and Zheng, Tianyu and Ni, Yuansheng and Wang, Yubo and Zhang, Kai and Tong, Shengbang and Sun, Yuxuan and Yu, Botao and Zhang, Ge and Sun, Huan and others},
  booktitle={Proceedings of the 63rd Annual Meeting of the Association for Computational Linguistics (Volume 1: Long Papers)},
  pages={15134--15186},
  year={2025}
}

@inproceedings{rezatofighi2019generalized,
  title={Generalized intersection over union: A metric and a loss for bounding box regression},
  author={Rezatofighi, Hamid and Tsoi, Nathan and Gwak, JunYoung and Sadeghian, Amir and Reid, Ian and Savarese, Silvio},
  booktitle={Proceedings of the IEEE/CVF conference on computer vision and pattern recognition},
  pages={658--666},
  year={2019}
}

@misc{LabelStudio,
  title={{Label Studio}: Data labeling software},
  url={https://github.com/HumanSignal/label-studio},
  note={Open source software available from https://github.com/HumanSignal/label-studio},
  author={
    Maxim Tkachenko and
    Mikhail Malyuk and
    Andrey Holmanyuk and
    Nikolai Liubimov},
  year={2020-2025},
}

@inproceedings{yao2025argus,
  title={Argus Inspection: Do Multimodal Large Language Models Possess the Eye of Panoptes?},
  author={Yao, Yang and Li, Lingyu and Song, Jiaxin and Chen, Chiyu and He, Zhenqi and Wang, Yixu and Wang, Xin and Gu, Tianle and Li, Jie and Teng, Yan and others},
  booktitle={Proceedings of the 33rd ACM International Conference on Multimedia},
  pages={13133--13140},
  year={2025}
}

@article{zhang2023unmanned,
  title={Unmanned aerial vehicle navigation in underground structure inspection: A review},
  author={Zhang, Ran and Hao, Guangbo and Zhang, Kong and Li, Zili},
  journal={Geological Journal},
  volume={58},
  number={6},
  pages={2454--2472},
  year={2023},
  publisher={Wiley Online Library}
}

@article{li2025verifybench,
  title={Verifybench: A systematic benchmark for evaluating reasoning verifiers across domains},
  author={Li, Xuzhao and Li, Xuchen and Hu, Shiyu and Guo, Yongzhen and Zhang, Wentao},
  journal={arXiv preprint arXiv:2507.09884},
  year={2025}
}

@article{li2025causalstep,
  title={Causalstep: A benchmark for explicit stepwise causal reasoning in videos},
  author={Li, Xuchen and Li, Xuzhao and Hu, Shiyu and Huang, Kaiqi and Zhang, Wentao},
  journal={arXiv preprint arXiv:2507.16878},
  year={2025}
}

@article{hu2024fiova,
  title={FIOVA: A Multi-Annotator Benchmark for Human-Aligned Video Captioning},
  author={Hu, Shiyu and Li, Xuchen and Li, Xuzhao and Zhang, Jing and Wang, Yipei and Zhao, Xin and Cheong, Kang Hao},
  journal={arXiv preprint arXiv:2410.15270},
  year={2024}
}

@article{li2025select,
  title={Select Less, Reason More: Prioritizing Evidence Purity for Video Reasoning},
  author={Li, Xuchen and Li, Xuzhao and Hu, Shiyu and Huang, Kaiqi},
  journal={arXiv preprint arXiv:2510.15440},
  year={2025}
}

@article{cao2025large,
  title={Large language models for planning: A comprehensive and systematic survey},
  author={Cao, Pengfei and Men, Tianyi and Liu, Wencan and Zhang, Jingwen and Li, Xuzhao and Lin, Xixun and Sui, Dianbo and Cao, Yanan and Liu, Kang and Zhao, Jun},
  journal={arXiv preprint arXiv:2505.19683},
  year={2025}
}

@article{li2025sciagent,
  title={SciAgent: A Unified Multi-Agent System for Generalistic Scientific Reasoning},
  author={Li, Xuchen and Wu, Ruitao and Liu, Xuanbo and Wang, Xukai and Hu, Jinbo and Bai, Zhixin and Zeng, Bohan and Liang, Hao and Chen, Leheng and Chen, Mingrui and others},
  journal={arXiv preprint arXiv:2511.08151},
  year={2025}
}

@inproceedings{li2025multimodal,
  title={Multimodal knowledge retrieval-augmented iterative alignment for satellite commonsense conversation},
  author={Li, Qian and Li, Xuchen and Chang, Zongyu and Zhang, Yuzheng and Ji, Cheng and Wang, Shangguang},
  booktitle={Proceedings of the Thirty-Fourth International Joint Conference on Artificial Intelligence},
  pages={8168--8176},
  year={2025}
}

@inproceedings{li2024dtllm,
  title={Dtllm-vlt: Diverse text generation for visual language tracking based on llm},
  author={Li, Xuchen and Feng, Xiaokun and Hu, Shiyu and Wu, Meiqi and Zhang, Dailing and Zhang, Jing and Huang, Kaiqi},
  booktitle={Proceedings of the IEEE/CVF Conference on Computer Vision and Pattern Recognition},
  pages={7283--7292},
  year={2024}
}
\clearpage
\appendix
\renewcommand{\thetable}{A\arabic{table}}
\renewcommand{\thefigure}{A\arabic{figure}}
\setcounter{figure}{0}  
\setcounter{table}{0}  

\section{Annotation and Data Collection}
\begin{figure}[t!]
    \centering
    \includegraphics[width=0.48\textwidth]{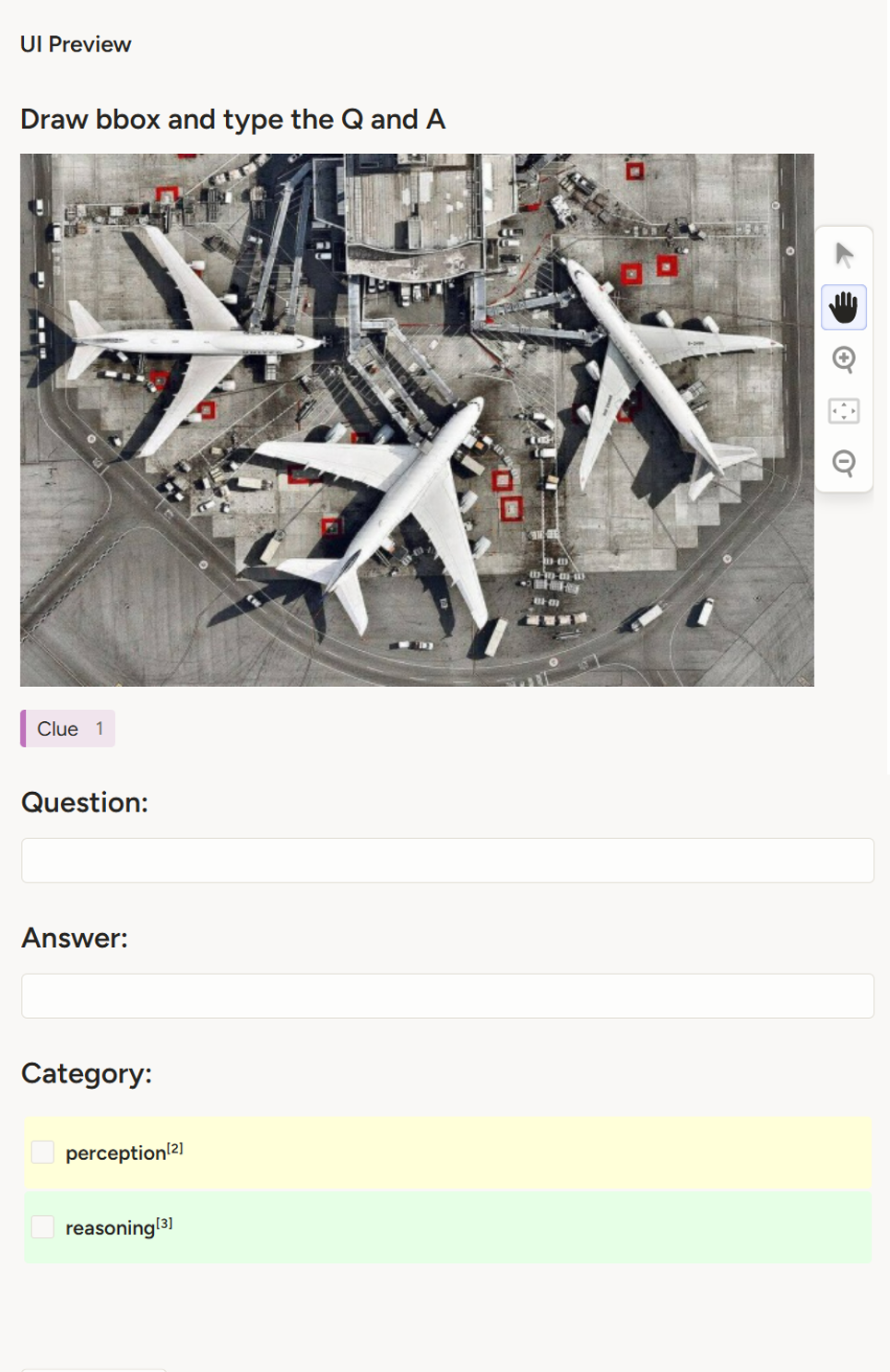}
    \caption{The web-based annotation interface used for ViEBench data collection. It supports interactive bounding box drawing, Q\&A entry, and fine-grained category selection.}
    \label{fig:annotation_interface}
\end{figure}

To ensure high-quality grounding and reasoning pairs, we developed a specialized web-based annotation platform based on Label Studio \cite{LabelStudio} as shown in Fig. \ref{fig:annotation_interface}. The interface is designed to facilitate the synchronized collection of visual bounding boxes, natural language queries, and logical categorizations.

The annotation pipeline consists of the following four key modules:
\begin{itemize}[itemsep=2pt, topsep=0pt, parsep=2pt]
    \item \textbf{Interactive Image Canvas}: The core of the interface allows annotators to perform multi-level inspection. Supporting standard zooming and panning operations, it enables annotators to locate minuscule targets that require agentic ``Thinking-with-Images'' behavior.
    \item \textbf{Bounding Box Grounding}: Annotators are required to draw a precise gold bounding box ($B_{gt}$) around the visual evidence necessary to answer the question. This provides the ground-truth for calculating the Intersection-over-Area (IoA) metrics used in our evaluation.
    \item \textbf{Q\&A Annotation}: Two dedicated text fields are provided for annotators to author the Question and the corresponding Answer. Annotators are instructed to ensure that the question cannot be answered confidently without referring to the fine-grained details within the specified bounding box.
    \item \textbf{Task Categorization}: Each sample is manually classified into one of two categories:
    \begin{enumerate}
        \item \textbf{Perception}: Questions focusing on direct attribute recognition or simple object identification.
        \item \textbf{Reasoning}: Questions requiring multi-step logical deduction, spatial relationship analysis, or the synthesis of internal knowledge with visual evidence.
    \end{enumerate}
\end{itemize}

\section{Case Study}
\begin{figure}[t!]
    \centering
    \includegraphics[width=0.48\textwidth]{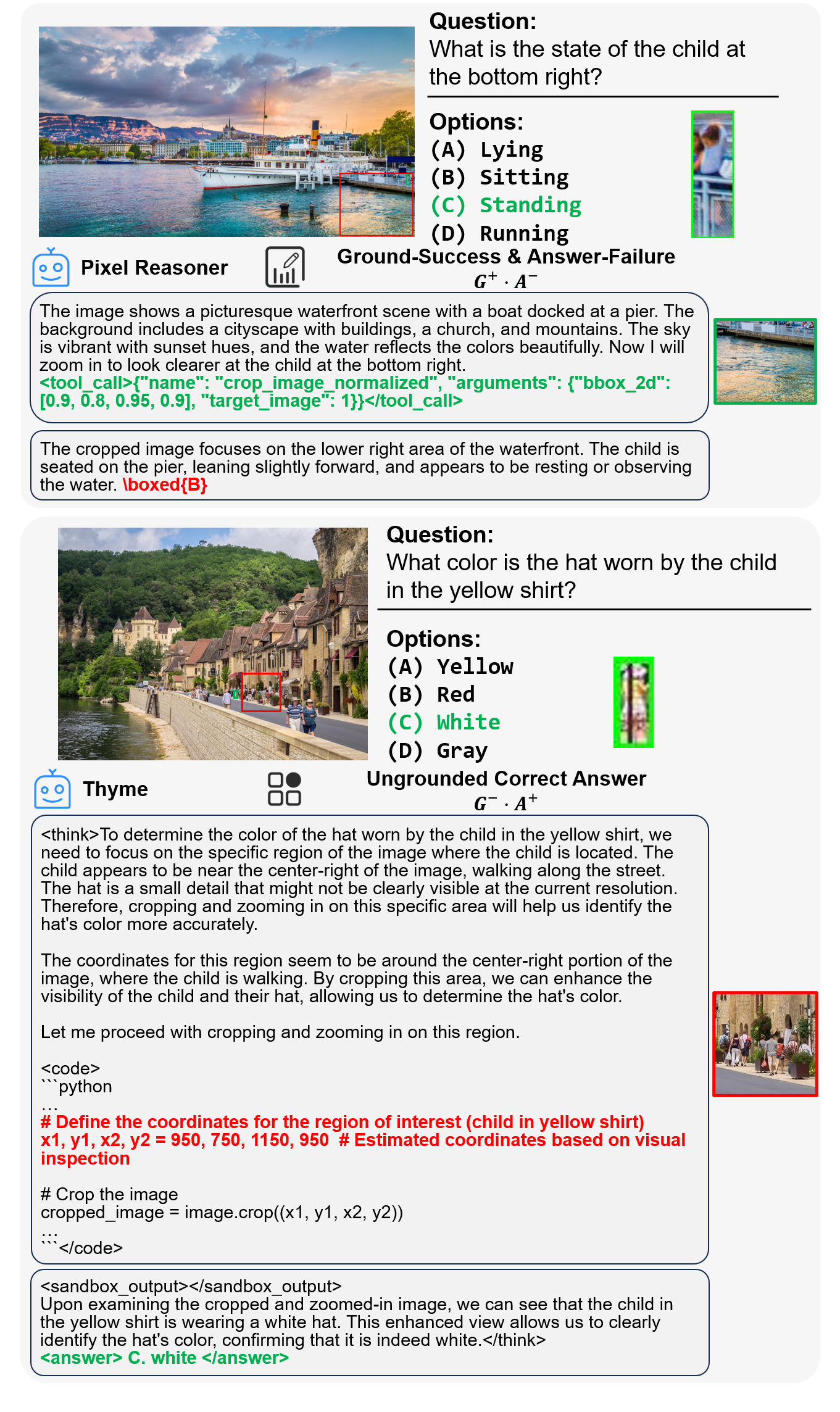}
    \caption{(Top) Ground-Success and Answer-Failure: Pixel Reasoner achieves near-perfect spatial grounding on the target child but fails to correctly interpret the child's physical state from the high-resolution crop, leading to an incorrect final answer. (Bottom) Ungrounded Correct Answer: Thyme arrives at the correct answer despite focusing on an irrelevant background region far from the gold BBox. This exposes a redundant cropping behavior, where the model's tool-invocation process is functionally decoupled from its final decision.}
    \label{fig:visualization}
\end{figure}

In our case studies (Fig. \ref{fig:visualization}), Pixel Reasoner \cite{pixelreasoner} demonstrates a representative instance of Ground-Success \& Answer-Failure ($G^+ \cdot A^-$). When tasked with a query requiring specific posture recognition, the model accurately identifies the need to focus on the "child at the bottom right" and generates a high-precision crop that perfectly encompasses the gold BBox. However, it still misinterprets the child's posture as "sitting" rather than "standing." This instance serves as evidence that successful visual localization does not inherently guarantee logical understanding. It highlights a cognitive bottleneck where models struggle to synthesize fine-grained visual semantics into a correct reasoning chain, even when the relevant pixels are clearly in view.

Conversely, Thyme \cite{thyme} illustrates Ungrounded Correct Answer ($G^- \cdot A^+$) behavior when tasked to determine the "color of the hat worn by the child in the yellow shirt." Although its CoT correctly identifies the intent to zoom into the specific region, the actual executed crop coordinates are significantly shifted toward an irrelevant background area, missing the gold BBox entirely. Despite this localization failure, the model arrives at the correct answer. This phenomenon suggests that the model may be utilizing broader global context cues rather than specific grounded evidence. Such cases underscore the necessity of process-level auditing provided by ViEBench to distinguish between faithful visual operations and results that lack grounded evidence.

\end{document}